\pdfoutput=1

\documentclass[11pt]{article}
\newcommand{\sys}{\textsc{JointLM}}
\newcommand{\sysseknow}{\textsc{SeKnow}}
\newcommand{\syssep}{\textsc{SepLM}}
\newcommand{\tfidf}{\textsc{TF-IDF}}
\newcommand{\datasetstruct}{\textsc{SeKnow-MultiWOZ}}
\newcommand{\datasetsemi}{\textsc{HybridToD}}
\newcommand{\datasetunstruct}{\textsc{UnstructuredToD}}
\usepackage[]{EACL2023}

\usepackage{times}
\usepackage{latexsym}
\usepackage{algorithm}
\usepackage{algorithmic}
\usepackage{amsmath}
\usepackage{subcaption}
\usepackage{paralist}
\usepackage{multirow}
\usepackage{comment}
\usepackage{svg}
\usepackage{array}
\usepackage{xcolor}

\usepackage{newfloat}
\usepackage{listings}
\usepackage{booktabs}
\usepackage{todonotes}
\usepackage[T1]{fontenc}

\usepackage[utf8]{inputenc}

\usepackage{microtype}

\usepackage{inconsolata}

\title{Joint Reasoning on Hybrid-knowledge sources for Task-Oriented Dialog}

\author{Mayank Mishra \\
  IBM Research \\
  \texttt{mayank.mishra1@ibm.com} \\\And
  Danish Contractor \\
  IBM Research \\
  \texttt{danish.contractor@ibm.com} \\\And
  Dinesh Raghu \\
  IBM Research \\
  \texttt{diraghu1@in.ibm.com} \\}

\begin{document}
\maketitle
\begin{abstract}
Traditional systems designed for task oriented dialog utilize knowledge present only in structured knowledge sources to generate responses. However, relevant information required to generate responses may also reside in unstructured sources, such as documents. Recent state of the art models such as HyKnow \cite{HyKnow} and \sysseknow\ \cite{SeKnow} aimed at overcoming these challenges make limiting assumptions about the knowledge sources. For instance, these systems assume that certain types of information, such as a phone number, is {\em always} present in a structured knowledge base (KB) while information about aspects such as entrance ticket prices, would always be available in documents.

In this paper, we create a modified version of the MutliWOZ-based dataset prepared by \cite{SeKnow} to demonstrate how current methods have significant degradation in performance when strict assumptions about the source of information are removed.
Then, in line with recent work exploiting pre-trained language models, we fine-tune a BART \cite{BART} based model using prompts \cite{NEURIPS2020_1457c0d6,DBLP:journals/corr/abs-2107-02137} for the tasks of querying knowledge sources, as well as, for response generation, without making assumptions about the information present in each knowledge source. Through a series of experiments, we demonstrate that our model is robust to perturbations to knowledge modality (source of information), and that it can fuse information from structured as well as unstructured knowledge to generate responses.
\end{abstract}

\section{Introduction}
Most existing work on task-oriented dialog systems assumes that the knowledge required for completing a task (eg: booking a restaurant reservation), resides in structured knowledge sources. Thus, typical task-oriented dialog systems require generating a {\em belief state}, that can be used to query a knowledge base to fetch entity results; these results are then used to generate responses. Recognizing that information is not always present in structured resources, recently methods that can additionally use unstructured knowledge (eg: document collections), have also been developed \cite{DSTC9, SeKnow}. However, current state-of-the-art models designed for such tasks make limiting assumptions about the nature of knowledge sources, that make them unsuitable for use in real-world settings. 


\noindent{\bf Limitations of existing methods:}  First, current task-oriented dialog systems designed to reason over hybrid knowledge sources assume that a knowledge base and the unstructured knowledge source encode separate pieces of information about entities (eg: the zip-code is always in structured knowledge, ticket prices are always available in unstructured text) \cite{DSTC9, assumption}. This is not reflective of real-world knowledge, where independent information systems are often fused to enable applications. 

Second, existing systems are trained to learn the source of different pieces of information, thus, making them  unsuitable for situations where any field that was previously in a structured knowledge source is now available in an unstructured knowledge source (and vice versa). In effect, a simple change in the modality of information can result in a failure of the model to utilize the information present in knowledge, as existing models memorize the source of every piece of information.

\begin{figure*}[ht]
    \centering
    \includegraphics[scale=0.75]{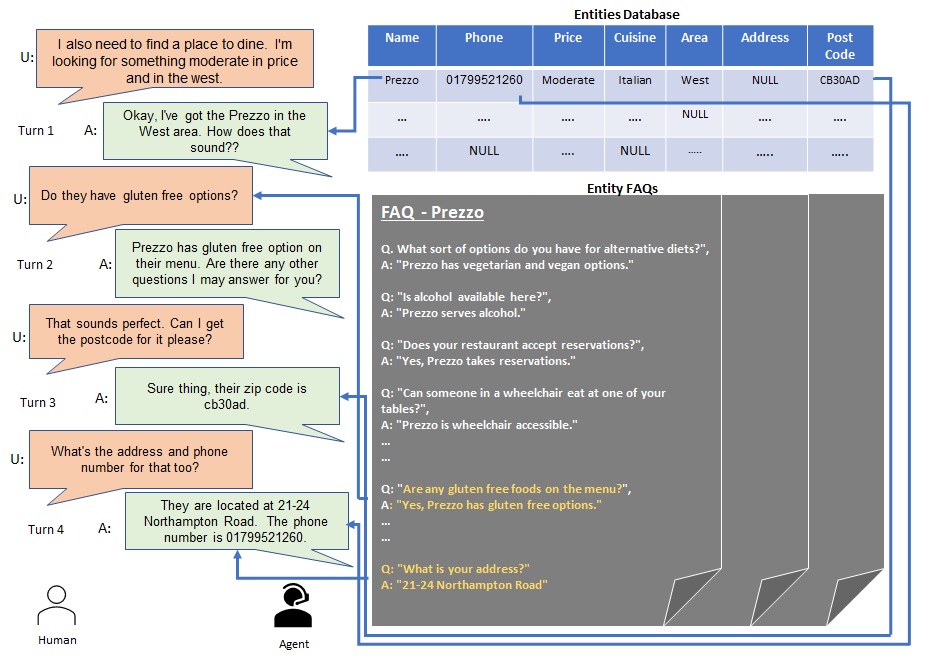}
    \caption{Example of a dialog requiring the use of data from two different sources. Agent Turn 4, requires incorporating information from both structured (DB Table) and unstructured data (a document consisting of FAQs for the entity). }
    \label{fig:main_example}
\end{figure*}

Third, such systems assume that each knowledge grounded response can contain information from only {\em one} source type \cite{DSTC9,SeKnow,assumption} -- either structured or unstructured knowledge. This is an artificial constraint imposed to make modelling easier, but real-world conversations can routinely require systems to fuse information from more than one knowledge type (eg: See Dialog turn $4$ in Figure \ref{fig:main_example}).   

\noindent{\bf Contributions: } In this paper, we present our work aimed at removing each of these strict assumptions from task-oriented dialog systems. Current methods for joint-reasoning in task oriented dialogs have been developed using an augmented version of MultiWOZ 2.1 which contains additional dialog turns based on new unstructured information \cite{SeKnow}. Unfortunately, no attempt has been made to distribute information across knowledge sources. 
We therefore create a modified version of this dataset (called \datasetsemi) that optimally redistributes information across structured and unstructured knowledge so that most dialogs in the train dataset are affected by this change.

A trivial method of redistributing information across structured and unstructured knowledge sources would be to arbitrarily move structured fields for some entities to the unstructured knowledge source. However, since the universe of entities in the dataset is very large and not all entities are directly referred to in the dialogs, such a method of redistributing information may not be as effective if the dialogs do not use the slot-values that have been redistributed. We therefore, develop an automated graph based approach which uses the max-cut of the graph to optimally redistribute information from structured to unstructured knowledge sources. 


\begin{table*}
\center 
\scriptsize
\begin{tabular}{llll}
\toprule

\textbf{Slot Type} & \textbf{Slot Values} & \textbf{Question Template}           & \textbf{Answer Template}                            \\
\midrule
price              & cheap                & What is the price range?             & It has \$\{price\} pricing.                         \\
                   & expensive            & How costly is \$\{restaurant name\}? & \$\{restaurant name\} is \$\{price\}                    \\
\midrule
cuisine            & Italian              & What is the cuisine?                 & \$\{restaurant name\} caters for \$\{cuisine\} cuisine. \\
                   & Thai                 & What type of food is served here?    & You can find \$\{cuisine\} food here  \\
\bottomrule
\end{tabular}
\caption{Examples of templates used for moving slot values from the structured to the unstructured knowledge source.}
\label{table:templates}
\end{table*}
Lastly, in line with recent work exploiting pre-trained language models, we fine-tune BART \cite{BART} using prompts for the tasks of querying knowledge as well as response generation without making assumptions about the information present in each knowledge source. Specifically, we do Prompt+LM finetuning \cite{DBLP:journals/corr/abs-2107-13586} in which both the prompt and model parameters are trainable \cite{DBLP:journals/corr/abs-2102-12206,DBLP:journals/corr/abs-2103-10385,DBLP:journals/corr/abs-2105-11259}. Through a series of experiments, we demonstrate that our model is robust to perturbations to knowledge modality (source of information), and it can fuse information from structured as well as unstructured knowledge to generate responses. 

In summary we make the following contributions\footnote{The constructed dataset and code used is available at \href{https://github.com/mayank31398/HybridToD}{https://github.com/mayank31398/HybridToD}}:
    (1) We prepare a new version of the MultiWOZ-DSTC9 combined dataset \cite{DSTC9,SeKnow} called \datasetsemi~ to study the reasoning on hybrid knowledge sources for task oriented dialog systems.
    (2) We demonstrate that our model (referred to as \sys) is also able to fuse information from both knowledge modalities and beats existing state-of-the-art systems on standardized metrics.
    (3) We present detailed ablation studies demonstrating the value of our modelling choices.

\section{Related Work}
\noindent{\bf Modeling Task Oriented Dialogs:} Multiple flavours of this problem have been defined to address different aspects of modeling - eg: belief state tracking to assess whether a model is able to correctly decode the query needed given a current conversational context ~\cite{BST01,BST02,yang2021ubar}, generating responses given belief states to assess whether a model is able to correctly predict the knowledge attributes to be used in a response \cite{yang2021ubar,chen2019working,gao-etal-2020-paraphrase,mohapatra-etal-2021-simulated-chats}, end-to-end modeling of dialog systems where models are assessed on the correctness of the response generated including the values used from the knowledge base \cite{bordes2017learning,raghu-etal-2021-constraint}, etc. Recent work that assumes that belief state annotations are latent and not available for training have also been developed \cite{10.1162/tacl_a_00372}.

\noindent{\bf Knowledge Grounded Dialog:} Dialog systems that generate responses on information grounded in external knowledge have also been developed. Unlike, work on task oriented dialogs, which primarily focuses on using structured knowledge to complete a `goal' or accomplish a `task' (eg: POI recommendation for in car navigation \cite{eric-etal-2017-key}, restaurant, hotel or flight booking \cite{el-asri-etal-2017-frames}, etc), most existing knowledge grounded systems are designed to address informational needs of users (eg: answering queries based on collections of documents, making response recommendations to contact center agents). 
Finally, contemporaneous to our work, knowledge grounded response generation tasks that combine information from hybrid knowledge sources have also been proposed  \cite{nakamura-etal-2022-hybridialogue}. Here, unlike task oriented dialog systems, which require the retrieval of an entity to make recommendations or accomplish a task, in such tasks, the goal is to answer an informational seeking query in a chit-chat conversation. Models are required to use the dialog context to fetch related tables (often flattened and encoded as independent table cells), along with documents to generate a response.  

\section{The \datasetsemi\ Dataset}\label{dataset}
The dataset prepared by \cite{HyKnow} (referred to as the \datasetstruct ~dataset in this paper) is the only publicly available task-oriented dialog dataset in which the dialogs are grounded on two types of knowledge sources: structured and unstructured (FAQs). However, \datasetstruct\ is not indicative of a real-world setting due to two major limitations: (1) It has a strict, slot-type to knowledge-source type mapping. For example, the slot-type `cuisine' is always in the structured source while `timings' of operation would always be mentioned in unstructured documents, and (2) an agent response contains information from only one source (i.e., either from structured or unstructured). To alleviate these limitations, we systematically modify the knowledge sources in \datasetstruct\ to construct a new dataset that we refer to as \datasetsemi.


\noindent{\bf Dataset Construction}: 
We first create an undirected graph $G = (V, E)$ where each vertex $v \in V$ is a unique slot-value and an edge $e \in E$ exists between two vertices, if the slot values represented by these vertices occur together in a training dialog utterance. For instance, in Figure \ref{fig:main_example} nodes associated with slot-values ``{\em 21-24 Northampton Road}'' and phone number ``{\em 01799521660}'' would have an edge between them due to Turn 4. Similarly, vertices corresponding to the values for slot-type `cuisine' \textit{Italian} and the slot-type `address' \textit{21-24 Northampton Road} would have had an edge between them if the utterance at Turn 4 was instead, ``\textit{It is an {Italian} restaurant located at {21-24 Northampton Road}}". Our goal is to move some of the slot-values (vertices) in $G$ that are originally in the structured knowledge source to the unstructured knowledge source so as to alleviate some of the limitations of the original dataset.

In order to identify which vertices to move, we create a maxcut of the graph $G$ using the MaxCutBM algorithm \cite{MaxCut}. A maxcut results in a graph in which the most number of edges from the original graph are `cut'. 
After the application of MaxCut, all slot values in one graph partition are retained in the structured knowledge source, while the others are converted to text QA pairs  using templates and included as part of the unstructured document associated with that entity. 
The templates for the restaurant slot types `price' and `cuisine' are shown in Table \ref{table:templates} for illustration. Since the edges of the graph are based on slot-value mentions in dialog utterances, applying a maxcut modifies the knowledge source in a way that it affects most dialog turns in the dataset; in other words, the max-cut ensures the maximum possible utterances in the dataset have information fused from both knowledge sources.
\begin{table}
\tiny
\begin{center}
\begin{tabular}{|c|c|c|c|c|}
\hline

& \multicolumn{3}{c|}{\bf Context-Response pairs} & {\bf Number of entities}
\\
\hline
\textbf{Domain} & train & validation & test & train/validation/test
\\
\hline
hotel & 19370 & 2316 & 2295 & 33
\\
restaurant & 19716 & 2162 & 2188 & 110
\\
attraction & 8192 & 1226 & 1246 & 79
\\
\hline
total & 47278 & 5704 & 5729 & 222
\\
\hline

\end{tabular}
\caption{Number of context-response pairs in the dataset}
\label{table:context-response}
\end{center}
\end{table}




\begin{table}
\tiny
\begin{center}
\begin{tabular}{|c|c|c|c|}
\hline

{\bf Domain} & \datasetstruct & \datasetsemi 
\\
\hline
hotel & 10.97 & 6.79 
\\
restaurant & 8.12 & 5.25 
\\
attraction & 9 & 6.38 
\\
\hline

\end{tabular}
\caption{Average number of slot values by domain in the structured knowledge source for each dataset.}
\label{table:structured}
\end{center}
\end{table}

\begin{table}
\tiny
\begin{center}
\begin{tabular}{|c|c|c|c|}
\hline

{\bf Domain} & \datasetstruct & \datasetsemi & \datasetunstruct
\\
\hline
hotel & 36.52 & 40.58 & 46.48
\\
restaurant & 14.96 & 17.83 & 22.7
\\
attraction & 0 & 2.62 & 8
\\
\hline
\end{tabular}
\caption{Average number of FAQs for each domain in the unstructured knowledge source. }
\label{table:unstructured}
\end{center}
\end{table}

Since we move slot values from one partition of the graph to the unstructured knowledge source, a slot type can now have some values in structured knowledge and some in unstructured knowledge (as an FAQ). We find that our approach ends up modifying each entity referred to in the dataset, and that slot-values of the same type are now distributed across different types of knowledge. 

For experimentation, we also create a version of the dataset with all slot values\footnote{The entity name is also a slot type but we always retain in it in both knowledge sources.} moved from the structured to the unstructured knowledge source. We refer to this dataset as \datasetunstruct. To construct \datasetsemi\ and \datasetunstruct\ dataset, we only consider dialogs from 3 domains: hotel, restaurant and attraction. We omit dialogs from other domains as they do not have associated knowledge. For example, the taxi domain only contains the information that the slot-type \textit{phone} should match the regular expression \verb/["^[0-9]{10}$"]/, but does not contain any instance of phone numbers present in the train dialogs.


\noindent{\bf Dataset Statistics:}
The number of context-response pairs (spread across the 3 domains: hotel, restaurant and attractions) for \datasetsemi\ are shown in Table \ref{table:context-response}. We also show the entity distribution by domain-type. The restaurant domain dominates the knowledge sources, occupying almost half of the total entities and the other half is constituted by hotel and attraction domains.
Tables \ref{table:structured} and \ref{table:unstructured} show the distribution of entity slot-values in structured knowledge sources and FAQs in the unstructured knowledge source for each domain in the datasets. As can be seen, the average number of slot-values presented in structured knowledge are lesser in \datasetsemi\ as compared to \datasetstruct\, and correspondingly the number of FAQs in \datasetsemi\ are higher as compared to \datasetstruct.  We  present the detailed slot-type distribution of \datasetstruct\ and \datasetsemi\ in the appendix. 
We find that approximately 50\% slot-values are moved to unstructured knowledge from the structured sources for each slot-type. 

\noindent{\bf Limitations of the Dataset:} Information about entities is only redistributed from the structured knowledge source to the unstructured knowledge source. In effect, information that was previously in unstructured knowledge sources continues to remain there. Redistributing information from unstructured documents to structured documents would require annotations to be able to extract facets to be converted to slot-types. 

We describe our model, \sys\ in the next section.

\section{\sys}

The problem of utilizing information and responding to users in task-oriented dialogs can be broken down into parts: (i) Querying Knowledge Source (structured and/or unstructured) to return entities (ii) Generating Responses (eg:sharing information about entities, requesting for more details from the user, etc). 

\begin{table*}[ht]
\scriptsize
\begin{center}
\begin{tabular}{|c|c|c|c|c|c|c|c|}
\hline
\multicolumn{5}{|c|}{} & \multicolumn{3}{c|}{\bf slot-values}
\\
\hline
{\bf Train Dataset} & {\bf Test Dataset} & {\bf Model} & {\bf Bleu-1} & {\bf Bleu-4} & {\bf prec.} & {\bf recall} & {\bf F1}
\\
\hline
\datasetsemi & \datasetstruct & \sys & \textbf{30.63} & \textbf{8.60} & \textbf{50.48} & \textbf{45.37} & \textbf{47.79}
\\
& & \sysseknow & 29.20 & 7.83 & 43.16 & 28.65 & 33.14
\\
\hline
\datasetsemi & \datasetsemi & \sys & \textbf{30.59} & \textbf{8.67} & \textbf{50.56} & \textbf{45.83} & \textbf{48.08}
\\
& & \sysseknow & 29.05 & 7.70 & 44.29 & 29.12 & 35.14
\\
\hline
\datasetsemi & \datasetunstruct & \sys & \textbf{30.30} & \textbf{8.44} & \textbf{51.05} & \textbf{45.37} & \textbf{48.04}
\\
& & \sysseknow & 27.43 & 6.68 & 42.96 & 19.62 & 27.11
\\
\hline
\end{tabular}
\caption{All models trained on \datasetsemi\ and evaluated on the rest of the datasets}
\label{table:hybridtod}
\end{center}
\end{table*}

\begin{table*}[ht]
\scriptsize
\begin{center}
\begin{tabular}{|c|c|c|c|c|c|c|c|}
\hline
\multicolumn{5}{|c|}{} & \multicolumn{3}{c|}{\bf slot-values}
\\
\hline
{\bf Train Dataset} & {\bf Test Dataset} & {\bf Model} & {\bf Bleu-1} & {\bf Bleu-4} & {\bf prec.} & {\bf recall} & {\bf F1}
\\
\hline
\datasetstruct & \datasetstruct & \sys & 29.07 & 8.06 & 49.74 & 41.31 & 45.13
\\
& & \sysseknow & \textbf{31.00} & \textbf{9.14} & \textbf{52.17} & \textbf{44.98} & \textbf{48.31}
\\
\hline
\datasetstruct & \datasetsemi & \sys & \textbf{27.77} & \textbf{7.54} & \textbf{44.48} & \textbf{36.39} & \textbf{40.03}
\\
& & \sysseknow & 26.61 & 7.32 & 42.19 & 26.70 & 33.31
\\
\hline
\datasetstruct & \datasetunstruct & \sys & \textbf{27.03} & \textbf{7.17} & \textbf{46.29} & \textbf{34.93} & \textbf{39.82}
\\
& & \sysseknow & 26.19 & 6.42 & 41.96 & 19.48 & 26.53
\\
\hline
\end{tabular}
\caption{All models trained on \datasetsemi\ and evaluated on the rest of the datasets}
\label{table:seknow}
\end{center}
\end{table*}
We represent the dialog context as $c = (u_1, r_1, ..., u_n)$, where $(u_i, r_i)$ represent the user and the system response utterance at $i^{th}$ turn respectively. We represent the entity $e$ required for generating the response as the concatenation of its slot-values (from structured KB), represented as $e^s$, and FAQs from the unstructured knowledge source, represented as $e^{us}$:

\begin{align*}
    \left[ e^s \right] = &\left< struct \right> \left< slot \right> slot_1 \left< val \right> value_1
    \\
    &\left< slot \right> slot_2 \left< val \right> value_2...
    \\
    \left[ e^{us} \right] = &\left< unstruct \right> \left< doc \right> document_1
    \\
    &\left< doc \right> document_2...
    \\
    \left[ e \right] =& \left[ e^s \right] \left[ e^{us} \right]
\end{align*}
where $\left< struct \right>$, $\left< unstruct \right>$ are special tokens to demarcate the start of structured knowledge and unstructured knowledge of an entity respectively. $\left< slot \right>$, $\left< val \right>$ demarcate the slot-type and its value and $\left< doc \right>$ denotes the start of a document from unstructured knowledge. We train \sys\ to jointly model two tasks: entity retrieval and response generation. We use a hyperparameter $\alpha$ to weigh the two tasks during training, where $\alpha$ denotes the number of training samples used for entity retrieval task. Note that $\alpha = 0.5$ denotes equal number of examples for both the tasks.

\subsection{Entity Retrieval}
As discussed, prior to generating a response, we need to retrieve the relevant entity required to generate the response. We represent the inputs to the language model (LM) for this task as:
\begin{align*}
    &\left< entity\_retrieval\_task \right> \left< u \right> u_1 \left< r \right> r_1 ...
    \\
    &\left< u \right> u_{n} \left< entity \right> [e_j]
\end{align*}
where, $e_j \in \mathcal{E}$, the set of all entities, $\left< entity\_retrieval\_task \right>$ and $\left< entity \right>$ are special tokens for task prompting and demarcating the start of an entity. We train the model to generate the special tokens $z_j = \left< relevant \right>$ or $z_j = \left< irrelevant \right>$ for each entity $e_j$ given the context $c$. We choose the best entity $e$ as:
\begin{equation}
    e = \underset{e_j}{\arg\!\max}\ \ p \left(z_j = \left< relevant \right> | c, e_j \right)
    \label{eq:inference}
\end{equation}

During training we use a subset of the entities in $\mathcal{E}$ for creating the positive and negative set of entities. However, at inference time, we evaluate on all the entities in $\mathcal{E}$.

\subsection{Response Generation}
After scoring all entities, we use the context and the best entity $e$ (the entity with the highest score for the $\left< relevant \right>$ token) and generate response using the same LM. We represent the inputs for this task as:
\begin{align*}
    &\left< response\_task \right> \left< u \right> u_1 \left< r \right> r_1 ...
    \\
    &\left< u \right> u_{n} \left< entity \right> [e]
\end{align*}

where $\left< response\_task \right>$ is a special token to prompt this task. 
We train the model to generate the response token-by-token.

\subsection{Training details}
We train our model to minimize $\sum_{(c, r)} \mathcal{L}(\theta, c, r)$, where
\begin{align*}
    \mathcal{L}(\theta, c, r) = &- \alpha \log p_\theta(z_j | c, e_j)\\
    &- (1 - \alpha) \log p_\theta(r | c, e_j)
\end{align*}
The first term in the above objective represents the log-likelihood of retrieving the relevant entity and the second term is the log-likelihood of generating the response. Note that the term $\alpha$ (percentage of samples for each task) can be adjusted by changing the number of examples for the two tasks in a given batch of fixed size.

To train our model, we use early stopping with $patience = 5$ for the above objective on the validation set to prevent overfitting of our model. The loss was optimized using AdamW optimizer \cite{AdamW}. We use a batch-size of 8 examples, with 4 examples for entity retrieval and 4 for response generation per batch. For the 4 examples for entity retrieval, 2 are positive and 2 are negative examples (effectively our batch is $2+2+4$). We use equation \ref{eq:inference} during inference to pick the highest scored relevant entity.

\section{Experiments}

\noindent Our experiments are aimed at answering the following questions:
(1) How does \sys\ perform compared to the baseline when trained and tested on \datasetsemi?
(2) How does the change in slot-value distribution across structured and unstructured sources affect the performance of the models?
(3) Is joint training of PromtLM for the two tasks of entity retrieval and response generation helpful?
(4) How does \sys\ compare with natural baselines for entity retrieval? 

\begin{table*}[ht]
\scriptsize
\begin{center}
\begin{tabular}{|c|c|c|c|c|c|c|c|c|c|}
\hline
{\bf Train Dataset} & {\bf Test Dataset} & {\bf Model} & {\bf success@1} & {\bf success@5} & {\bf Bleu-1} & {\bf Bleu-4} & {\bf prec.} & {\bf recall} & {\bf F1}
\\
\hline
& & \sys & \textbf{84.50} & \textbf{86.57} & \textbf{30.59} & \textbf{8.67} & \textbf{50.56} & \textbf{45.83} & \textbf{48.08}
\\
\datasetsemi & \datasetsemi & \syssep & 79.79 & 85.64 & 29.96 & 8.66 & 47.08 & 42.53 & 44.69
\\
& & \tfidf & 28.31 & 34.49 & - & - & - & - & -
\\
\hline
\end{tabular}
\caption{Performance of models on the entity retrieval task. }
\label{table:retrieval}
\end{center}
\end{table*}

\noindent {\bf Experimental Setup: }Task oriented dialog systems have to identify relevant entities (e.g. restaurants) from associated knowledge sources needed to generate a response. In order to identify these relevant entities, existing datasets provide the belief state annotations during training. 
Additionally, in our work for each dialog context, we associate  a set of (positive) entities that exactly match the requirements present in the dialog context and a set of (negative) entities that do not match by an automated method. Note that the text snippets in the unstructured corpus do not have any annotations.

For all of our experiments, we use BART \cite{BART} encoder-decoder based language model and finetune the pretrained model on the three datasets i.e, \datasetstruct\ \cite{SeKnow}, \datasetsemi\ and \datasetunstruct\ datasets. 

\noindent {\bf Baseline: }We use the current state-of-the-art model for joint reasoning,  \sysseknow\ \cite{SeKnow} model as our baseline. \sysseknow\ is designed to use belief state annotations 
-- specifically,  \sysseknow\ is trained to generate the belief state given the dialog context. These belief states are then used to query the knowledge sources and generate a delexicalised response using the context and the generated belief state. The slot-values in the delexicalised response are then populated using an unordered set of entities returned by the belief state query on the structured knowledge source.  

\subsection{Evaluation Metrics}
We report BLEU scores for assessing response generation performance and slot-value precision, recall and $F1$ for comparing the slot-value filling performance against the baseline. As described previously, since no new slot types were created from unstructured documents, the slot-value metrics are computed only using the slot-types that were originally present in the structured knowledge source. 

We also report success@k for entity retrieval baselines to assess the performance of systems on the entity selection task. We define success@k as $1$ if the top-k scored entities contain a relevant entity for response generation and $0$ otherwise. However, note that it is not possible to measure success@k on \sysseknow\ since it generates the response using an unordered set of entities returned by the belief state query. We thus compare the two models only based on their performance on response generation. 



\subsection{Results}\label{sec:results}
{\bf Knowledge-Source Memorization: }We train and test both \sys\ and the baseline model, \sysseknow\ on \datasetsemi\ and observe that \sys\ outperforms \sysseknow\ by $13$ points on slot-value F1 score (Row 1, Table \ref{table:hybridtod}). Also, the performance of \sysseknow\ drops from $48.31$ (Row 1, Table \ref{table:seknow}) when trained/tested on the \datasetstruct\ dataset to $35.14$ (Row 1, Table \ref{table:hybridtod}) when trained/tested on \datasetsemi ~dataset. This severe drop in performance is indicative of the fact that \sysseknow\ learns the source of slot-values and is unable to use information when the source of the particular slot-value can be varying (structured/unstructured) across entities.

\noindent{\bf Generalization of \sys}: To assess the generalization performance of the models, we train all the models on \datasetsemi\ and test on other datasets which have different slot-value distributions. As can be seen from Table \ref{table:hybridtod}, when trained on \datasetsemi, \sys\ outperforms \sysseknow\ on all three dataset settings, \datasetstruct, \datasetsemi\ and \datasetunstruct\ across all response generation metrics. We also notice that \sys\ trained on \datasetsemi\ is robust to change in the knowledge modality during inference (slot-value F1 stays at approx. $48$). This is not the case for \sysseknow\ which exhibits large drop (31\% from \datasetstruct\ to \datasetsemi\ and 45\% from \datasetstruct\ to \datasetunstruct) in slot-value F1, as the distribution of slot-types changes in different datasets (Table \ref{table:hybridtod}).

We also train the models on \datasetstruct, and test on the other datasets and notice that \sys\ outperforms \sysseknow\ on both \datasetsemi\ and \datasetunstruct\ (Table \ref{table:seknow}). However, \sysseknow\ has better slot-value F1 than \sys\ on \datasetsemi. We hypothesize that this is because the belief state labels are more informative and provide a very strong signal for \sysseknow\ on \datasetstruct\ and this has the effect of \sysseknow\ learning the knowledge modality which is not the case for \sys. This suggests that \sys\ has better generalization performance. 




\subsection{Model Ablation Study}

To study the importance of joint-training of our model, we also train a model without prompts using entity annotations, where two different BART \cite{BART} models are trained for retrieval and generation. We call this model \syssep. This model is trained on \datasetsemi\ and is compared against \sys\ on both entity retrieval and response generation (Table \ref{table:retrieval}). We observe that \sys\ outperforms \syssep\ in both the tasks with a 5 points difference in success@1 and a 3 points difference in slot-value F1. This confirms that the joint modeling of the 2 tasks using prompting yields a better model than learning a separate model for the 2 tasks at hand. For a detailed evaluation on all other dataset combinations, please refer to the Appendix. For comparison with a non-neural entity retrieval baseline, we also report the success scores BM25 based \tfidf\ retriever which are significantly worse than the neural retrievers used for \sys\ and \syssep. These experiments highlight the benefit of joint modeling of the two tasks.

\subsection{Qualitative Study} In Table \ref{table:dialog-example}, we show the responses generated for a sample dialog by \sys\ and \sysseknow\ on the three datasets used for our experiments. It should be noted that \sys\ generates the same response for all the three datasets. However, \sysseknow\, is not able to populate the required slot-values for this entity (Meze Bar) in the response in \datasetsemi\ and \datasetunstruct\ when those slot-values are no longer available in the structured source.
\begin{table}[]
\scriptsize
\begin{flushleft}
\textbf{\underline{Dialog context}}
\\
\textbf{User}: Suggest me some \underline{\textit{Turkish}} restaurants in Cambridge.
\\
\textbf{Agent}: Yes there is a \underline{\textit{Turkish}} restaurant in Cambridge with two different locations, would you like the addresses?
\\
\textbf{User}: I would like if its located in the \underline{\textit{center}} of the city price one located in the center of city and I don't care about the pricing.
\\
\textbf{\underline{Response}}
\\
\textbf{Ground truth response}: I have two \underline{\textit{Turkish}} restaurants, both in the \underline{\textit{centre}} and both \underline{\textit{expensive}}. May I recommend \underline{\textit{Meze Bar}}?
\end{flushleft}

\centering
\begin{tabular}{|c|c|c|}
\hline
&  \sysseknow & \sys\ \\ \hline
SeKnow-MultiWOZ & \begin{tabular}[c]{@{}l@{}}I found a restaurant at \underline{\textit{196}}\\ \underline{\textit{Mill Road City Centre}}. Their\\ phone number is\\ \underline{\textit{01223362372}}.\end{tabular}   & \multirow{3}{*}{\begin{tabular}[c]{@{}l@{}}\underline{\textit{Meze Bar}} is a \\ \underline{\textit{Turkish}} restaurant\\ in the \underline{\textit{expensive}}\\ price range. Their\\ address is \underline{\textit{196}}\\ \underline{\textit{Mill Road City}}\\ \underline{\textit{Centre}}.\end{tabular}}  \\  \cline{1-2}
HybridToD       &    \begin{tabular}[c]{@{}l@{}}I found a restaurant at \underline{\textit{196}}\\ \underline{\textit{Mill Road City Centre}}. Their\\ phone number is \{null\}.\end{tabular}                                                                                                                                                                            &     \\ \cline{1-2}  
UnstructuredToD &     \begin{tabular}[c]{@{}l@{}}I found a restaurant at \{null\}.\\ Their phone number is \{null\}.\end{tabular}                                                                                                                                                                             &                         \\ \hline
\end{tabular}
\caption{Response generated by \sysseknow\ and \sys\ on the three datasets.}
\label{table:dialog-example}
\end{table}

\section{Conclusion}
In this paper we presented a new dataset, \datasetsemi\ that requires reasoning over both structured and unstructured knowledge sources to generate responses to dialogs. Unlike existing task-oriented dialog datasets, it does not restrict slot-types to specific knowledge sources. Through our experiments we demonstrated how existing methods do not adapt well to changing distributions of slot-type sources and that our model \sys\ (trained using entity annotations rather than belief state), not only generates better responses by reasoning over both knowledge sources, it also learns a better retriever for entities. 
In future work, we also plan to train our models without using any annotations i.e without any supervision on entity label information.

\section{Limitations}
Our dataset and model are not intended to be directly used in a real-world system as they have some inherent limitations. As mentioned in Section \ref{dataset}, we only redistribute slot types from structured knowledge sources to unstructured knowledge sources.  Due to a lack of resources we are unable to annotate unstructured documents -- our dataset has a bias that certain information will always appear in unstructured information.  In addition, we rely on a pre-trained language model, BART, to generate responses. We have not assessed to what extent the generated responses could exhibit any form of social bias or toxic language (when prompted). We do not recommend that our system be used in a real-world deployed chatbot without further study. Lastly, this work has been assessed only on English language data using a pretrained language model developed for English. 

\section{Acknowledgement}
The authors would like to thank Raunak Sinha for his contributions on the scripts for the development of the dataset.

\bibliography{anthology}

\begin{thebibliography}{26}
\expandafter\ifx\csname natexlab\endcsname\relax\def\natexlab#1{#1}\fi

\bibitem[{Ben{-}David et~al.(2021)Ben{-}David, Oved, and
  Reichart}]{DBLP:journals/corr/abs-2102-12206}
Eyal Ben{-}David, Nadav Oved, and Roi Reichart. 2021.
\newblock \href {http://arxiv.org/abs/2102.12206} {{PADA:} {A} prompt-based
  autoregressive approach for adaptation to unseen domains}.
\newblock \emph{CoRR}, abs/2102.12206.

\bibitem[{Bordes et~al.(2017)Bordes, Boureau, and Weston}]{bordes2017learning}
Antoine Bordes, Y-Lan Boureau, and Jason Weston. 2017.
\newblock \href {https://openreview.net/forum?id=S1Bb3D5gg} {Learning
  end-to-end goal-oriented dialog}.
\newblock In \emph{International Conference on Learning Representations}.

\bibitem[{Boumal et~al.(2016)Boumal, Voroninski, and Bandeira}]{MaxCut}
Nicolas Boumal, Vladislav Voroninski, and Afonso~S. Bandeira. 2016.
\newblock The non-convex burer–monteiro approach works on smooth semidefinite
  programs.
\newblock In \emph{Proceedings of the 30th International Conference on Neural
  Information Processing Systems}, NIPS'16, page 2765–2773, Red Hook, NY,
  USA. Curran Associates Inc.

\bibitem[{Brown et~al.(2020)Brown, Mann, Ryder, Subbiah, Kaplan, Dhariwal,
  Neelakantan, Shyam, Sastry, Askell, Agarwal, Herbert-Voss, Krueger, Henighan,
  Child, Ramesh, Ziegler, Wu, Winter, Hesse, Chen, Sigler, Litwin, Gray, Chess,
  Clark, Berner, McCandlish, Radford, Sutskever, and
  Amodei}]{NEURIPS2020_1457c0d6}
Tom Brown, Benjamin Mann, Nick Ryder, Melanie Subbiah, Jared~D Kaplan, Prafulla
  Dhariwal, Arvind Neelakantan, Pranav Shyam, Girish Sastry, Amanda Askell,
  Sandhini Agarwal, Ariel Herbert-Voss, Gretchen Krueger, Tom Henighan, Rewon
  Child, Aditya Ramesh, Daniel Ziegler, Jeffrey Wu, Clemens Winter, Chris
  Hesse, Mark Chen, Eric Sigler, Mateusz Litwin, Scott Gray, Benjamin Chess,
  Jack Clark, Christopher Berner, Sam McCandlish, Alec Radford, Ilya Sutskever,
  and Dario Amodei. 2020.
\newblock \href
  {https://proceedings.neurips.cc/paper/2020/file/1457c0d6bfcb4967418bfb8ac142f64a-Paper.pdf}
  {Language models are few-shot learners}.
\newblock In \emph{Advances in Neural Information Processing Systems},
  volume~33, pages 1877--1901. Curran Associates, Inc.

\bibitem[{Chen et~al.(2019)Chen, Xu, and Xu}]{chen2019working}
Xiuyi Chen, Jiaming Xu, and Bo~Xu. 2019.
\newblock A working memory model for task-oriented dialog response generation.
\newblock In \emph{Proceedings of the 57th Annual Meeting of the Association
  for Computational Linguistics}, pages 2687--2693.

\bibitem[{Dey and Desarkar(2021)}]{BST01}
Suvodip Dey and Maunendra~Sankar Desarkar. 2021.
\newblock Hi-dst: A hierarchical approach for scalable and extensible dialogue
  state tracking.
\newblock In \emph{Proceedings of the 22nd Annual Meeting of the Special
  Interest Group on Discourse and Dialogue}, pages 218--227.

\bibitem[{El~Asri et~al.(2017)El~Asri, Schulz, Sharma, Zumer, Harris, Fine,
  Mehrotra, and Suleman}]{el-asri-etal-2017-frames}
Layla El~Asri, Hannes Schulz, Shikhar Sharma, Jeremie Zumer, Justin Harris,
  Emery Fine, Rahul Mehrotra, and Kaheer Suleman. 2017.
\newblock \href {https://doi.org/10.18653/v1/W17-5526} {{F}rames: a corpus for
  adding memory to goal-oriented dialogue systems}.
\newblock In \emph{Proceedings of the 18th Annual {SIG}dial Meeting on
  Discourse and Dialogue}, pages 207--219, Saarbr{\"u}cken, Germany.
  Association for Computational Linguistics.

\bibitem[{Eric et~al.(2017)Eric, Krishnan, Charette, and
  Manning}]{eric-etal-2017-key}
Mihail Eric, Lakshmi Krishnan, Francois Charette, and Christopher~D. Manning.
  2017.
\newblock \href {https://doi.org/10.18653/v1/W17-5506} {Key-value retrieval
  networks for task-oriented dialogue}.
\newblock In \emph{Proceedings of the 18th Annual {SIG}dial Meeting on
  Discourse and Dialogue}, pages 37--49, Saarbr{\"u}cken, Germany. Association
  for Computational Linguistics.

\bibitem[{Gao et~al.(2021{\natexlab{a}})Gao, Takanobu, and Huang}]{SeKnow}
Silin Gao, Ryuichi Takanobu, and Minlie Huang. 2021{\natexlab{a}}.
\newblock \href {http://arxiv.org/abs/2106.11796} {End-to-end task-oriented
  dialog modeling with semi-structured knowledge management}.
\newblock \emph{CoRR}, abs/2106.11796.

\bibitem[{Gao et~al.(2021{\natexlab{b}})Gao, Takanobu, Peng, Liu, and
  Huang}]{HyKnow}
Silin Gao, Ryuichi Takanobu, Wei Peng, Qun Liu, and Minlie Huang.
  2021{\natexlab{b}}.
\newblock \href {https://doi.org/10.18653/v1/2021.findings-acl.139}
  {{H}y{K}now: End-to-end task-oriented dialog modeling with hybrid knowledge
  management}.
\newblock In \emph{Findings of the Association for Computational Linguistics:
  ACL-IJCNLP 2021}, pages 1591--1602, Online. Association for Computational
  Linguistics.

\bibitem[{Gao et~al.(2020)Gao, Zhang, Ou, and Yu}]{gao-etal-2020-paraphrase}
Silin Gao, Yichi Zhang, Zhijian Ou, and Zhou Yu. 2020.
\newblock \href {https://doi.org/10.18653/v1/2020.acl-main.60} {Paraphrase
  augmented task-oriented dialog generation}.
\newblock In \emph{Proceedings of the 58th Annual Meeting of the Association
  for Computational Linguistics}, pages 639--649, Online. Association for
  Computational Linguistics.

\bibitem[{Han et~al.(2021)Han, Zhao, Ding, Liu, and
  Sun}]{DBLP:journals/corr/abs-2105-11259}
Xu~Han, Weilin Zhao, Ning Ding, Zhiyuan Liu, and Maosong Sun. 2021.
\newblock \href {http://arxiv.org/abs/2105.11259} {{PTR:} prompt tuning with
  rules for text classification}.
\newblock \emph{CoRR}, abs/2105.11259.

\bibitem[{Kim et~al.(2020)Kim, Eric, Gopalakrishnan, Hedayatnia, Liu, and
  Hakkani-Tur}]{DSTC9}
Seokhwan Kim, Mihail Eric, Karthik Gopalakrishnan, Behnam Hedayatnia, Yang Liu,
  and Dilek Hakkani-Tur. 2020.
\newblock \href {https://aclanthology.org/2020.sigdial-1.35} {Beyond domain
  {API}s: Task-oriented conversational modeling with unstructured knowledge
  access}.
\newblock In \emph{Proceedings of the 21th Annual Meeting of the Special
  Interest Group on Discourse and Dialogue}, pages 278--289, 1st virtual
  meeting. Association for Computational Linguistics.

\bibitem[{Lewis et~al.(2020)Lewis, Liu, Goyal, Ghazvininejad, Mohamed, Levy,
  Stoyanov, and Zettlemoyer}]{BART}
Mike Lewis, Yinhan Liu, Naman Goyal, Marjan Ghazvininejad, Abdelrahman Mohamed,
  Omer Levy, Veselin Stoyanov, and Luke Zettlemoyer. 2020.
\newblock \href {https://doi.org/10.18653/v1/2020.acl-main.703} {{BART}:
  Denoising sequence-to-sequence pre-training for natural language generation,
  translation, and comprehension}.
\newblock In \emph{Proceedings of the 58th Annual Meeting of the Association
  for Computational Linguistics}, pages 7871--7880, Online. Association for
  Computational Linguistics.

\bibitem[{Li et~al.(2021)Li, Cao, Sridhar, Zhu, Li, Hamza, and McAuley}]{BST02}
Shuyang Li, Jin Cao, Mukund Sridhar, Henghui Zhu, Shang-Wen Li, Wael Hamza, and
  Julian McAuley. 2021.
\newblock Zero-shot generalization in dialog state tracking through generative
  question answering.
\newblock In \emph{Proceedings of the 16th Conference of the European Chapter
  of the Association for Computational Linguistics: Main Volume}, pages
  1063--1074.

\bibitem[{Liu et~al.(2021{\natexlab{a}})Liu, Yuan, Fu, Jiang, Hayashi, and
  Neubig}]{DBLP:journals/corr/abs-2107-13586}
Pengfei Liu, Weizhe Yuan, Jinlan Fu, Zhengbao Jiang, Hiroaki Hayashi, and
  Graham Neubig. 2021{\natexlab{a}}.
\newblock \href {http://arxiv.org/abs/2107.13586} {Pre-train, prompt, and
  predict: {A} systematic survey of prompting methods in natural language
  processing}.
\newblock \emph{CoRR}, abs/2107.13586.

\bibitem[{Liu et~al.(2021{\natexlab{b}})Liu, Zheng, Du, Ding, Qian, Yang, and
  Tang}]{DBLP:journals/corr/abs-2103-10385}
Xiao Liu, Yanan Zheng, Zhengxiao Du, Ming Ding, Yujie Qian, Zhilin Yang, and
  Jie Tang. 2021{\natexlab{b}}.
\newblock \href {http://arxiv.org/abs/2103.10385} {{GPT} understands, too}.
\newblock \emph{CoRR}, abs/2103.10385.

\bibitem[{Loshchilov and Hutter(2017)}]{AdamW}
Ilya Loshchilov and Frank Hutter. 2017.
\newblock Decoupled weight decay regularization.
\newblock \emph{arXiv preprint arXiv:1711.05101}.

\bibitem[{Mohapatra et~al.(2021)Mohapatra, Pandey, Contractor, and
  Joshi}]{mohapatra-etal-2021-simulated-chats}
Biswesh Mohapatra, Gaurav Pandey, Danish Contractor, and Sachindra Joshi. 2021.
\newblock \href {https://doi.org/10.18653/v1/2021.findings-emnlp.103}
  {Simulated chats for building dialog systems: Learning to generate
  conversations from instructions}.
\newblock In \emph{Findings of the Association for Computational Linguistics:
  EMNLP 2021}, pages 1190--1203, Punta Cana, Dominican Republic. Association
  for Computational Linguistics.

\bibitem[{Nakamura et~al.(2022)Nakamura, Levy, Tuan, Chen, and
  Wang}]{nakamura-etal-2022-hybridialogue}
Kai Nakamura, Sharon Levy, Yi-Lin Tuan, Wenhu Chen, and William~Yang Wang.
  2022.
\newblock \href {https://doi.org/10.18653/v1/2022.findings-acl.41}
  {{H}ybri{D}ialogue: An information-seeking dialogue dataset grounded on
  tabular and textual data}.
\newblock In \emph{Findings of the Association for Computational Linguistics:
  ACL 2022}, pages 481--492, Dublin, Ireland. Association for Computational
  Linguistics.

\bibitem[{Raghu et~al.(2021{\natexlab{a}})Raghu, Gupta, and
  Mausam}]{10.1162/tacl_a_00372}
Dinesh Raghu, Nikhil Gupta, and Mausam. 2021{\natexlab{a}}.
\newblock \href {https://doi.org/10.1162/tacl_a_00372} {{Unsupervised Learning
  of KB Queries in Task-Oriented Dialogs}}.
\newblock \emph{Transactions of the Association for Computational Linguistics},
  9:374--390.

\bibitem[{Raghu et~al.(2021{\natexlab{b}})Raghu, Jain, {Mausam}, and
  Joshi}]{raghu-etal-2021-constraint}
Dinesh Raghu, Atishya Jain, {Mausam}, and Sachindra Joshi. 2021{\natexlab{b}}.
\newblock \href {https://doi.org/10.18653/v1/2021.findings-acl.448} {Constraint
  based knowledge base distillation in end-to-end task oriented dialogs}.
\newblock In \emph{Findings of the Association for Computational Linguistics:
  ACL-IJCNLP 2021}, pages 5051--5061, Online. Association for Computational
  Linguistics.

\bibitem[{Sun et~al.(2021)Sun, Wang, Feng, Ding, Pang, Shang, Liu, Chen, Zhao,
  Lu, Liu, Wu, Gong, Liang, Shang, Sun, Liu, Ouyang, Yu, Tian, Wu, and
  Wang}]{DBLP:journals/corr/abs-2107-02137}
Yu~Sun, Shuohuan Wang, Shikun Feng, Siyu Ding, Chao Pang, Junyuan Shang,
  Jiaxiang Liu, Xuyi Chen, Yanbin Zhao, Yuxiang Lu, Weixin Liu, Zhihua Wu,
  Weibao Gong, Jianzhong Liang, Zhizhou Shang, Peng Sun, Wei Liu, Xuan Ouyang,
  Dianhai Yu, Hao Tian, Hua Wu, and Haifeng Wang. 2021.
\newblock \href {http://arxiv.org/abs/2107.02137} {{ERNIE} 3.0: Large-scale
  knowledge enhanced pre-training for language understanding and generation}.
\newblock \emph{CoRR}, abs/2107.02137.

\bibitem[{Wolf et~al.(2020)Wolf, Debut, Sanh, Chaumond, Delangue, Moi, Cistac,
  Rault, Louf, Funtowicz, Davison, Shleifer, von Platen, Ma, Jernite, Plu, Xu,
  Scao, Gugger, Drame, Lhoest, and Rush}]{wolf-etal-2020-transformers}
Thomas Wolf, Lysandre Debut, Victor Sanh, Julien Chaumond, Clement Delangue,
  Anthony Moi, Pierric Cistac, Tim Rault, Rémi Louf, Morgan Funtowicz, Joe
  Davison, Sam Shleifer, Patrick von Platen, Clara Ma, Yacine Jernite, Julien
  Plu, Canwen Xu, Teven~Le Scao, Sylvain Gugger, Mariama Drame, Quentin Lhoest,
  and Alexander~M. Rush. 2020.
\newblock \href {https://www.aclweb.org/anthology/2020.emnlp-demos.6}
  {Transformers: State-of-the-art natural language processing}.
\newblock In \emph{Proceedings of the 2020 Conference on Empirical Methods in
  Natural Language Processing: System Demonstrations}, pages 38--45, Online.
  Association for Computational Linguistics.

\bibitem[{Yang et~al.(2021)Yang, Li, and Quan}]{yang2021ubar}
Yunyi Yang, Yunhao Li, and Xiaojun Quan. 2021.
\newblock Ubar: Towards fully end-to-end task-oriented dialog system with
  gpt-2.
\newblock In \emph{Proceedings of the AAAI Conference on Artificial
  Intelligence}, volume~35, pages 14230--14238.

\bibitem[{Zhang et~al.(2021)Zhang, Chen, Wu, Wan, and Li}]{assumption}
Weijie Zhang, Jiaoxuan Chen, Haipang Wu, Sanhui Wan, and Gongfeng Li. 2021.
\newblock \href {http://arxiv.org/abs/2106.14444} {A knowledge-grounded dialog
  system based on pre-trained language models}.
\newblock \emph{CoRR}, abs/2106.14444.

\end{thebibliography}
\bibliographystyle{acl_natbib}

\appendix

\section{Appendix}
\begin{figure*}
    \centering
    \begin{subfigure}[b]{0.495\textwidth}
        \centering
        \includegraphics[width=\textwidth]{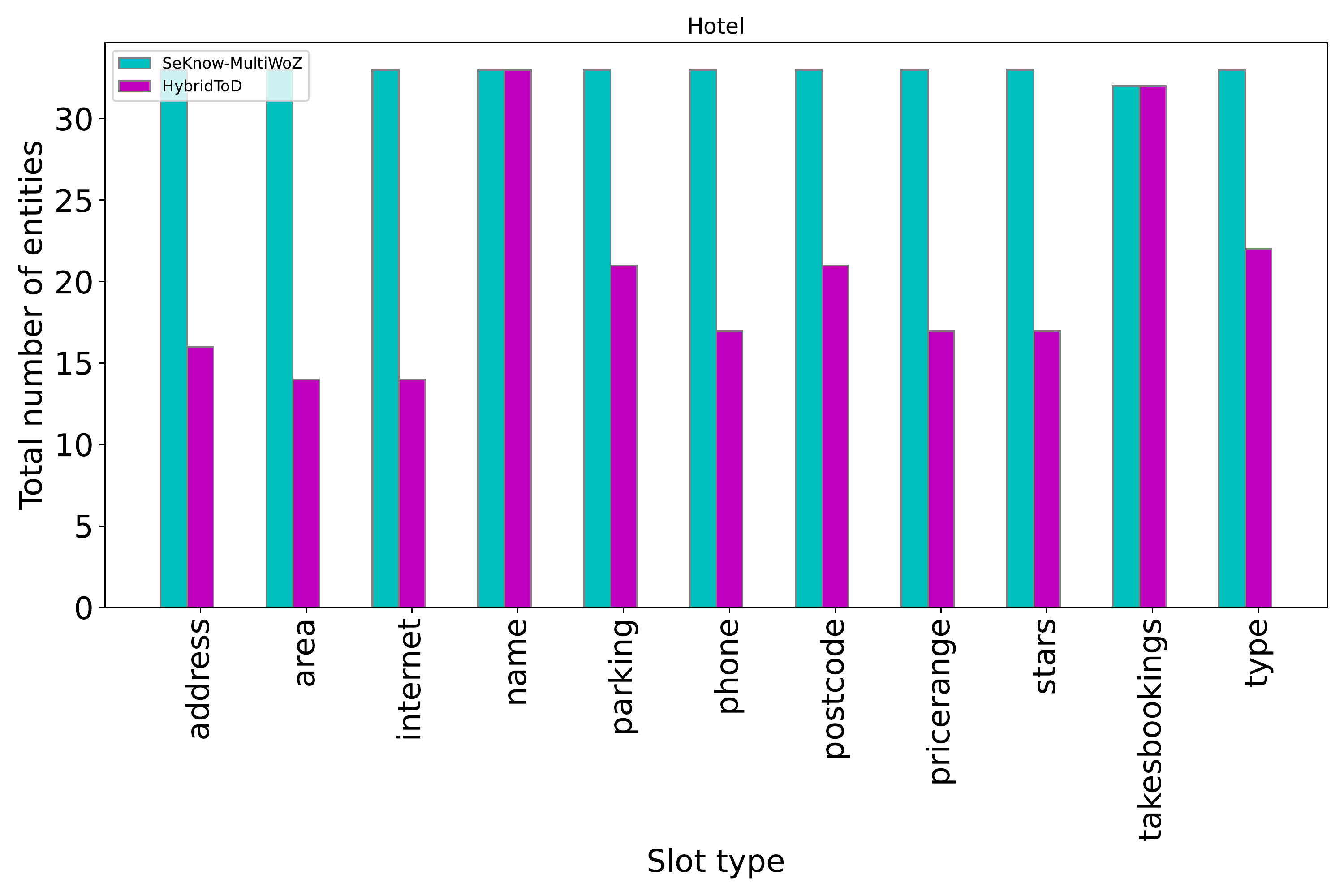}
        \caption{Hotel}
        \label{fig:slot-value-hotel}
    \end{subfigure}
    \hfill
    \begin{subfigure}[b]{0.495\textwidth}
        \centering
        \includegraphics[width=\textwidth]{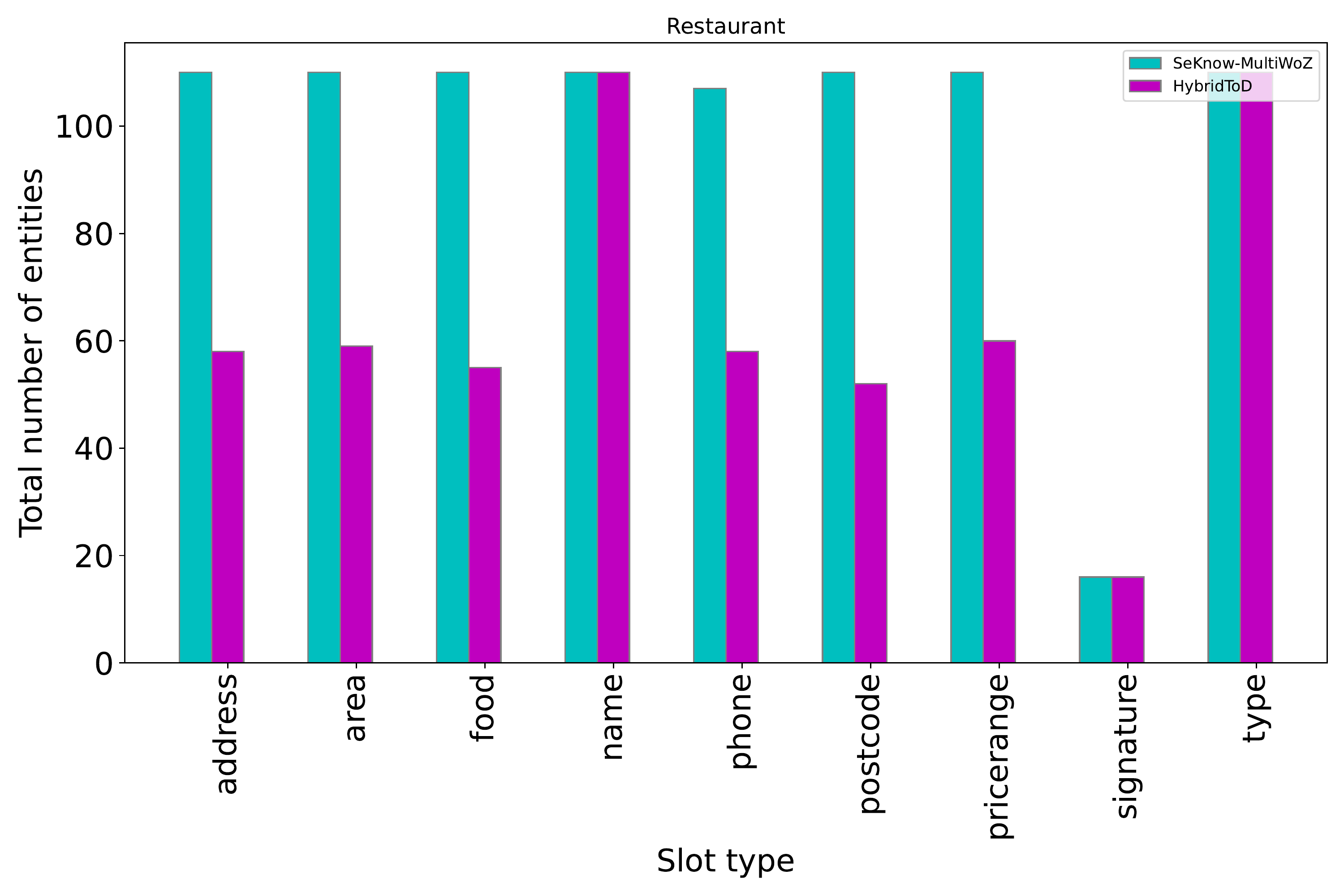}
        \caption{Restaurant}
        \label{fig:slot-value-restaurant}
    \end{subfigure}
    \caption{Figures \ref{fig:slot-value-hotel} and \ref{fig:slot-value-restaurant} show the slot-value distribution by slot-types in the hotel and restaurant domains the three datasets.}
    \label{fig:slot-value-distribution1} 
\end{figure*}

\begin{figure}
    \centering
        \includegraphics[scale=0.27]{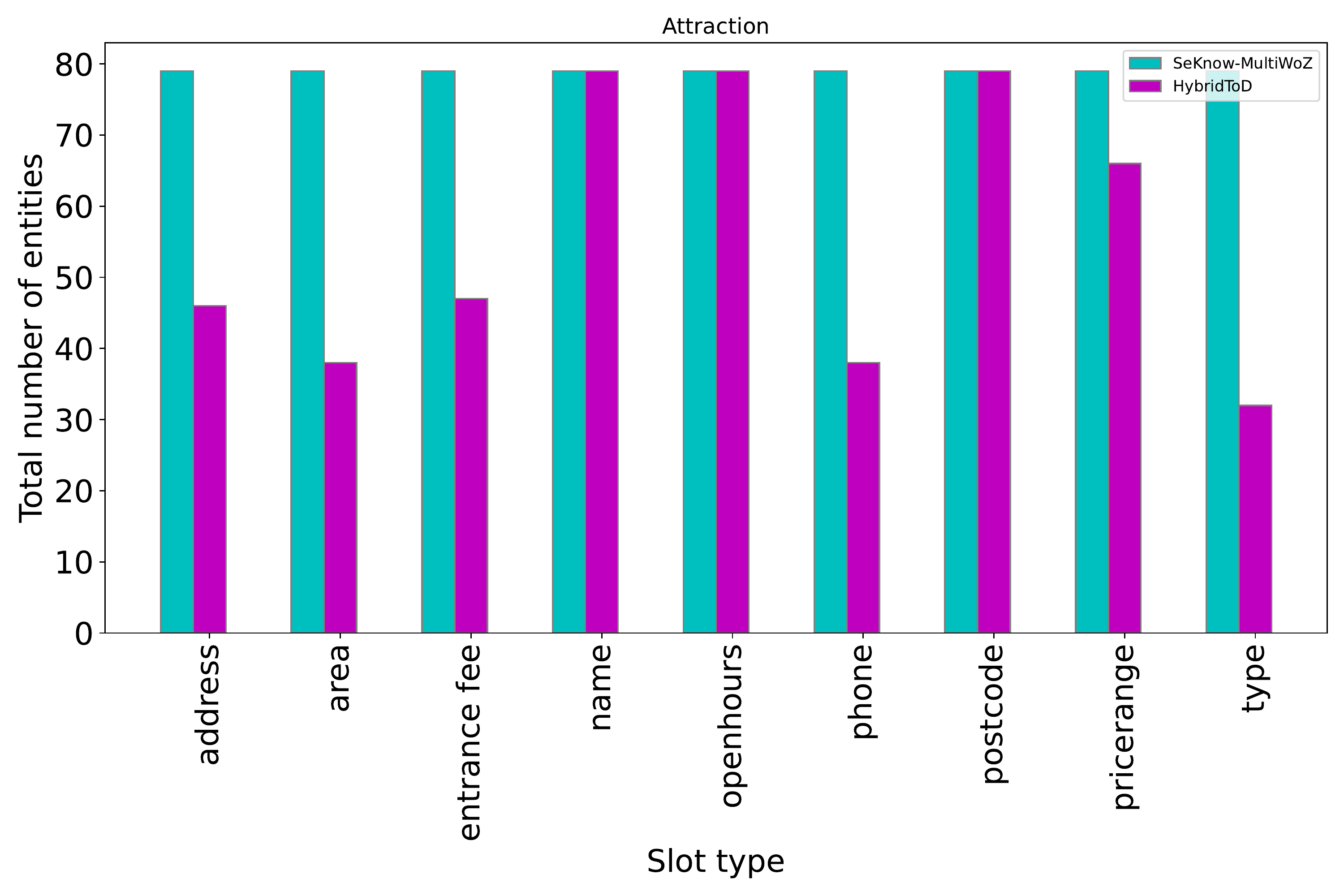}
    \caption{This figure shows the slot-value distribution by slot-types in the attraction domains in the three datasets.} 
    \label{fig:slot-value-distribution2}
\end{figure}

\begin{table*}
\scriptsize
\begin{center}
\begin{tabular}{|c|c|c|c|c|c|c|c|}
\hline
\multicolumn{5}{|c|}{} & \multicolumn{3}{c|}{\bf slot-values}
\\
\hline
{\bf Train Dataset} & {\bf Test Dataset} & {\bf Model} & {\bf Bleu-1} & {\bf Bleu-4} & {\bf prec.} & {\bf recall} & {\bf F1}
\\
\hline
& & \sys & \textbf{30.63} & 8.60 & \textbf{50.48} & \textbf{45.37} & \textbf{47.79}
\\
\datasetsemi & \datasetstruct & \syssep & 30.03 & \textbf{8.63} & 47.26 & 42.76 & 44.89
\\
& & \sysseknow & 29.20 & 7.83 & 43.16 & 28.65 & 33.14
\\
\hline
& & \sys & \textbf{30.59} & \textbf{8.67} & \textbf{50.56} & \textbf{45.83} & \textbf{48.08}
\\
\datasetsemi & \datasetsemi & \syssep & 29.96 & 8.66 & 47.08 & 42.53 & 44.69
\\
& & \sysseknow & 29.05 & 7.70 & 44.29 & 29.12 & 35.14
\\
\hline
& & \sys & \textbf{30.30} & \textbf{8.44} & \textbf{51.05} & \textbf{45.37} & \textbf{48.04}
\\
\datasetsemi & \datasetunstruct & \syssep & 29.78 & 8.41 & 47.08 & 41.63 & 44.19
\\
& & \sysseknow & 27.43 & 6.68 & 42.96 & 19.62 & 27.11
\\
\hline
\end{tabular}
\caption{All models trained on \datasetsemi\ and evaluated on the rest of the datasets}
\label{table:hybridtod-all}
\end{center}
\end{table*}

\subsection{Additional Results}
We present additional results for the comparison of \sys, \syssep\ and \sysseknow\ \cite{SeKnow} when trained on \datasetsemi\ and tested on the other datasets (Table \ref{table:hybridtod-all}). We see that \sys\ outperforms \syssep\ and \sysseknow\ on all the datasets demonstrating the importance of joint modeling.

\subsection{Additional Dataset Statistics}

We  present the detailed slot-type distribution of \datasetstruct\ and \datasetsemi\ in Figure \ref{fig:slot-value-distribution1} and \ref{fig:slot-value-distribution2}. 
We find that approximately 50\% slot-values are moved to unstructured knowledge from the structured sources for each slot-type. The bar-graphs show the number of entities with a particular slot-type.

\subsection{Hyperparameters and Training Details}
For all our experiments, we use BART \cite{BART} model from the HuggingFace Transformers library \cite{wolf-etal-2020-transformers}. To train the BART model, we use early stopping with $patience = 5$ on the validation set to prevent overfitting of both the entity retriever and the response generator. We use learning rate  = $10^{-5}$ with AdamW optimizer \cite{AdamW}. We use a batch-size of 8 examples, with 4 examples for entity retrieval and 4 for response generation per batch. For the 4 examples for entity retrieval, 2 are positive and 2 are negative examples (effectively our batch is $2+2+4$). All the experiments are conducted on a single A100 80GB GPU.





\end{document}